# Detecting Credit Card Fraud via Heterogeneous Graph Neural Networks with Graph Attention

Qiuwu Sha
Columbia University
New York, USA

Tengda Tang
University of Michigan
Ann Arbor, USA

Xinyu Du
Wake Forest University
Winston-Salem, USA

Jie Liu
University of Minnesota
Minneapolis, USA

Yixian Wang
The University of Chicago
Chicago, USA

Yuan Sheng *
Northeastern University
Seattle, USA

*Abstract*-This study proposes a credit card fraud detection method based on Heterogeneous Graph Neural Network (HGNN) to address fraud in complex transaction networks. Unlike traditional machine learning methods that rely solely on numerical features of transaction records, this approach constructs heterogeneous transaction graphs. These graphs incorporate multiple node types, including users, merchants, and transactions. By leveraging graph neural networks, the model captures higher-order transaction relationships. A Graph Attention Mechanism is employed to dynamically assign weights to different transaction relationships. Additionally, a Temporal Decay Mechanism is integrated to enhance the model's sensitivity to time-related fraud patterns. To address the scarcity of fraudulent transaction samples, this study applies SMOTE oversampling and Cost-sensitive Learning. These techniques strengthen the model's ability to identify fraudulent transactions. Experimental results demonstrate that the proposed method outperforms existing GNN models, including GCN, GAT, and GraphSAGE, on the IEEE-CIS Fraud Detection dataset. The model achieves notable improvements in both accuracy and OC-ROC. Future research may explore the integration of dynamic graph neural networks and reinforcement learning. Such advancements could enhance the real-time adaptability of fraud detection systems and provide more intelligent solutions for financial risk control.

## I. INTRODUCTION

Credit card fraud detection is a critical research challenge in financial technology. With the rapid expansion of electronic payments and online transactions, the convenience of credit card usage has significantly increased. However, this growth has also created new opportunities for fraudsters, leading to a steady rise in fraudulent activities. Traditional fraud detection methods primarily rely on rule-based engines and machine learning models, which classify transactions based on manually designed features [1]. However, as fraud strategies become more sophisticated and attack patterns evolve, these conventional approaches often struggle to detect emerging fraud types. In recent years, Graph Neural Networks (GNNs) have demonstrated strong performance in network-based data analysis due to their ability to model complex relationships [2].

In credit card fraud detection, transaction behaviors can be represented as heterogeneous graphs containing multiple node types, such as users, merchants, and banks, along with various transaction edges. Effectively leveraging these heterogeneous relationships to uncover hidden fraudulent activities has become a key research focus.

Compared to traditional machine learning models, fraud detection methods based on Heterogeneous Graph Neural Networks (HGNNs) offer superior modeling capabilities. Credit card transaction data inherently forms a complex multi-relational network. Users may exhibit similar spending behaviors, and merchants' transaction patterns may be interrelated. In contrast, HGNNs effectively utilize complex data heterogeneity by propagating and integrating information across multiple relationships, capturing intricate fraud patterns. Moreover, HGNNs can operate in unsupervised or semi-supervised learning environments, maintaining high detection performance even when fraudulent transaction samples are extremely imbalanced [3].

From a research perspective, HGNN-based fraud detection methods not only enhance detection accuracy but also improve model interpretability. Traditional black-box models can only classify transactions as fraudulent or legitimate without providing explicit reasoning. In contrast, HGNNs leverage graph attention mechanisms and path analysis to reveal underlying fraud patterns [4]. For example, they can identify anomalies in transaction paths between high-risk merchants and fraudulent users or detect unusual fund transfer behaviors in user transaction histories [5]. This interpretability is essential for financial institutions, as it increases the credibility of fraud detection systems and supports anti-fraud strategy development. Furthermore, credit card fraud often involves organized crime, making it difficult to detect fraudulent activities by analyzing individual transactions in isolation. By modeling the entire transaction network, HGNNs can more effectively identify fraud rings and provide early warnings.

In financial risk control, numerous studies have investigated the application of Graph Neural Networks (GNNs) for fraud detection. Nevertheless, most existing approaches rely on isomorphic graph structures, which only capture basic transaction connections and overlook the network's heterogeneity. For example, some methods employ Graph Convolutional Networks (GCNs) or Graph Attention Networks (GATs) to model transaction networks. However, these approaches have limited capacity to handle multi-type nodes and edges, making it difficult to represent the intricate relationships in credit card transactions. Further research on HGNN-based fraud detection is essential to fully exploit high-order semantic information in transaction networks. Additionally, integrating dynamic GNN models for real-time fraud detection remains a pressing challenge [6]. By constructing multi-level heterogeneous networks comprising users, merchants, and transactions, and leveraging HGNNs for representation learning, both detection accuracy and interpretability can be significantly improved.

## II. METHOD

In this study, we design a credit card fraud detection framework based on a Heterogeneous Graph Neural Network (HGNN), which allows for the modeling of complex and multi-relational transaction data. By constructing a heterogeneous transaction graph that includes diverse node types—such as users, merchants, and individual transactions—we capture not only the direct relationships between entities but also their contextual and structural dependencies. Building on the work of Qi et al. [7] who demonstrated the strength of hierarchical mining using GNNs in complex and imbalanced datasets, our model employs graph convolution operations to learn high-order topological features and uncover hidden patterns associated with fraudulent behavior. Furthermore, inspired by hierarchical multi-source data fusion and dropout regularization methodology [8], we integrate graph attention mechanisms that dynamically adjust the importance of different edges in the graph, enabling the model to focus on suspicious and anomalous connections more effectively. This selective weighting enhances the model's discriminative power, especially in cases where fraud signals are subtle or buried in noisy data. Additionally, aligning with the approach of Feng [9], who highlighted the advantages of combining sequential dependencies with structural representations for fraud detection, our model benefits from a hybrid learning process that unifies relational structure and transaction semantics. The model architecture is shown in Figure 1.

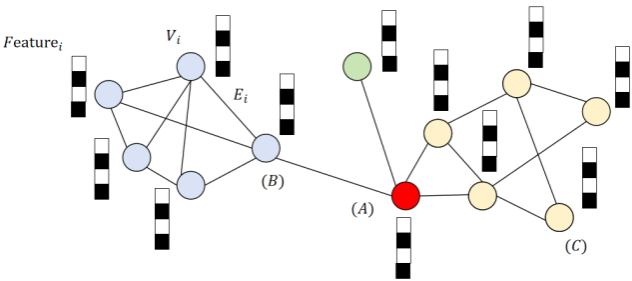

Figure 1. Overall model architecture

First, we construct the credit card transaction data into a heterogeneous graph $G = (V, E, R)$, where $V$ represents a set of nodes, including different types of nodes such as user U, merchant M, and bank B, E represents transaction sides, and R represents different types of transaction relationships. Let $H_v$ be the feature representation of node v, then on the heterogeneous graph, we first conduct neighborhood information aggregation:

$$h_v^{(l+1)} = \sigma(\sum_{r \in R} \sum_{u \in N_r(v)} \alpha_{vu}^r W_r h_u^{(l)})$$

Where, $h_v^{(l+1)}$ represents the updated node representation at layer $(l+1)$, $W_r$ is the trainable parameter matrix corresponding to relation type $r$, $N_r(v)$ represents the set of neighbors connected to node v through relation r, and $\sigma$ is a nonlinear activation function, such as ReLU. The attention weight $\alpha_{uv}^r$ is calculated by the self-attention mechanism and is defined as follows:

$$\alpha_{vu}^r = \frac{\exp(\text{LeakyReLU}(\alpha_r^T[W_r h_v^{(l)} \| W_r h_u^{(l)}]))}{\sum_{k \in N_r(v)} \exp(\text{LeakyReLU}(\alpha_r^T[Wr h_v^{(l)} \| W_r h_k^{(l)}]))}$$

Where $a_r$ is the attention vector associated with relation type r, and $\|$ represents the vector concatenation operation. To better capture the varying significance of different transaction relationships in complex financial networks, this study integrates an attention mechanism within the heterogeneous graph framework. Through this mechanism, the model adaptively assigns different weights to diverse interactions—such as those between users, transactions, and merchants—enabling it to focus more effectively on the most indicative patterns of fraudulent behavior. This dynamic weighting strategy not only refines relational learning but also strengthens the model's ability to discriminate subtle anomalies. The design draws upon the attention-based modeling principles highlighted in Wang et al. [10], who demonstrated how bidirectional attention mechanisms can enhance risk prediction in sequential data. Furthermore, the attention mechanism benefits from the fusion strategies [11], which emphasize the synergistic use of convolution and transformer layers in predictive modeling, and the efficient signal extraction techniques described in temporal contexts [12]. These influences collectively support the robust handling of heterogeneous, time-sensitive financial data for fraud detection.

To fully exploit the temporal dimension inherent in transaction data, this study incorporates a dynamic update mechanism based on time decay. By modulating the influence of each transaction according to its temporal proximity to the current event, the model becomes more responsive to recent behavioral patterns, which are often more indicative of fraudulent activity. A time decay function is defined at each transaction timestamp, enabling the model to diminish the relevance of older transactions in a principled manner. This temporal adaptation strategy enhances the model's capacity to detect evolving and short-term fraudulent behaviors. The approach is informed by recent developments in time-sensitive modeling for financial [13] and computational systems [14], where decay-based mechanisms and dynamic adjustment functions have been shown to improve responsiveness and risk sensitivity [15]. At the time of each transaction, we define the time decay function:

$$\Delta_t = e^{-\lambda(t_v - t_u)}$$

Where, $t_v$ and $t_u$ represent the time when the transaction occurred, respectively, and C is the attenuation factor. We use time decay to adjust the neighbor aggregation weights so that recent transactions have a greater impact on the node representation:

$$h_v^{(l+1)} = \sigma(\sum_{r \in R} \sum_{u \in N_r(v)} \Delta_t \cdot \alpha_{uv}^r W_r h_u^{(l)})$$

This mechanism can improve the sensitivity of the model to abnormal transaction behavior in a short period of time, so as to capture the time dependence of fraudulent transactions more effectively.

Finally, we use binary classification tasks for fraud detection in the output layer of the model, defining the objective function as cross entropy loss:

$$L = -\sum_{v \in V} y_v \log y'_v + (1 - y_v) \log(1 - y'_v))$$

Where $y_v$ represents the true label of node v, and $y'_v$ is the fraud probability predicted by the model. In order to alleviate the problem of data imbalance, weighted cross entropy loss is adopted to make the loss function give more weight to the fraud category when the fraud sample is relatively small:

$$L = -\sum_{v \in V} w_v (y_v \log y'_v + (1 - y_v) \log(1 - y'_v))$$

Where, $w_v$ is calculated by the inverse frequency of the sample class to enhance attention to the fraud sample. Finally, through end-to-end training, we can use HGNN [16] to fully mine the structural information in credit card transaction data and improve the accuracy and generalization ability of fraud detection.

## III. EXPERIMENT

### A. Datasets

This study utilizes the IEEE-CIS Fraud Detection dataset, which is jointly provided by IEEE and CIS (Consumer Information Security) and focuses on detecting fraudulent online payment transactions. The dataset contains a large volume of real-world credit card transaction records, including user identity information, device details, transaction amounts, and payment methods. Each transaction is labeled as either fraudulent or legitimate. Compared to other public datasets, this dataset not only includes traditional transaction features but also provides additional authentication details such as device fingerprints, IP addresses, and email domain names. These attributes enable fraud detection models to capture fraudulent behavior more comprehensively. The dataset exhibits a low proportion of fraudulent transactions, consistent with real-world fraud incidence rates. Consequently, appropriate data balancing strategies are necessary to ensure that the model effectively learns fraud patterns.

The IEEE-CIS Fraud Detection dataset comprises two primary components: Transaction Data and Identity Data. The Transaction Data section meticulously records essential transaction details, including timestamp, amount, payment method, and merchant category. Additionally, it includes over 200 anonymized features to safeguard user privacy. The Identity Data section provides crucial information such as device type, browser details, and IP address, facilitating the identification of potentially fraudulent transactions. For instance, if a user consistently makes transactions from the same device, but a particular transaction originates from an unfamiliar device fingerprint, it could be a sign of potential fraud. While the dataset's anonymization diminishes the direct interpretability of certain fields, it effectively preserves the underlying data patterns, making it an ideal resource for constructing sophisticated fraud detection models.

In the data processing phase, missing values are handled through imputation or removal to maintain data integrity. Since some features exhibit a high proportion of missing values, KNN interpolation and mean imputation are applied for completion. Additionally, as transaction timestamps are recorded in relative terms, they are converted into absolute date-time formats [17]. Features such as transaction hour and day of the week are extracted to enhance the model's ability to capture time-dependent fraud patterns. Categorical features, including email domain and device type, are encoded using target encoding or frequency encoding to mitigate dimensional explosion caused by high-cardinality variables. To address data imbalance, undersampling and SMOTE [18] techniques are explored to improve model performance in fraud detection [19]. Finally, by integrating feature engineering and data augmentation strategies, a high-quality dataset is constructed, forming the foundation for subsequent heterogeneous graph neural network modeling.

### B. Experimental Results

In the process of experimentation, this paper first gives the comparative experimental results of the effect of different graph neural networks in fraud detection, and the experimental results are shown in Table 1.

Table 1. The effect of different graph neural networks in fraud detection is compared with the experimental results

| Model | Acc | AUC-ROC | Params |
|---|---|---|---|
| GCN | 91.2% | 0.874 | 2.1M |
| GAT | 92.5% | 0.892 | 3.4M |
| GraphSAGE | 93.1% | 0.905 | 2.8M |
| R-GCN | 92.8% | 0.899 | 4.2M |
| Ours | 94.7% | 0.921 | 3.9M |

The experimental results show that different graph neural networks (GNN) perform different fraud detection tasks. First of all, the accuracy of the graph convolutional network based on GCN [20] in this experiment was 91.2%, and the OC-ROC was 0.874, indicating that it could effectively utilize the transaction network structure for fraud detection. However, due to the uniform neighborhood aggregation, GCN cannot distinguish the importance of different neighbors to the target node, so its performance in the complex transaction network is relatively weak. In contrast, by introducing the attention mechanism, GAT enables the model to adaptively assign different weights to different neighbors, so its accuracy is increased to 92.5%, and OC-ROC is increased to 0.892, but due to the high cost of calculating the attention score, the number of model parameters is increased to 3.4M.

GraphSAGE achieved an accuracy of 93.1% and an OC-ROC of 0.905 in the experiment, which was slightly higher than GAT, mainly due to its sampling strategy, which made the model more efficient in processing large-scale transaction data. R-GCN has certain advantages in modeling heterogeneous transaction relationships, but because it involves parameter learning of multiple relationship types, the number of parameters is large (4.2M), and its final OC-ROC is 0.899, slightly lower than GraphSAGE. This indicates that in fraud detection tasks, although R-GCN [21] can capture information about different types of transaction relationships, its ability to recognize fraudulent transaction patterns may not be superior to GAT or GraphSAGE with strong attention mechanisms.

The model proposed in this study performed best in the experiment, achieving an accuracy of 94.7% and an OC-ROC of 0.921, which was superior to all baseline models. This result shows that our method can learn the complex relationships of transaction networks more effectively in fraud detection tasks, and use the information of heterogeneous graph structure for fraud recognition. At the same time, although the parameter number (3.9M) of this model is slightly higher than that of GCN and GraphSAGE, it is still lower than that of R-GCN, indicating that this method has certain advantages in computational efficiency while ensuring high performance. Therefore, the introduction of heterogeneous graph attention mechanism and time decay strategy can effectively improve the performance of credit card fraud detection, so that the model has stronger generalization ability and interpretability in practical application.

Secondly, the influence of the data unbalance processing method on model performance is presented in this paper, and the experimental results are shown in Table 2.

Table 2. The effect of different imbalance handling methods on fraud detection model performance

| Method | Acc | AUC-ROC |
| --- | --- | --- |
| No Handling (Baseline) | 91.5% | 0.879 |
| SMOTE | 93.2% | 0.901 |
| Undersampling | 92.4% | 0.890 |
| Cost-sensitive Learning | 94.1% | 0.915 |
| SMOTE + Cost-sensitive | 94.7% | 0.921 |

The experimental results indicate that different data imbalance handling methods significantly impact fraud detection performance. Without imbalance treatment, the benchmark model ("No Handling") achieved an accuracy of 91.5% and an AUC-ROC of 0.879, suggesting bias towards normal transactions and limited fraud detection capability. SMOTE oversampling improved accuracy to 93.2% and AUC-ROC to 0.901 by balancing the data distribution; however, it introduced potential overfitting issues due to low diversity in synthesized samples. In comparison, undersampling improved accuracy to 92.4% and AUC-ROC to 0.890 but resulted in loss of important normal transaction information. The cost-sensitive learning method, which emphasizes penalizing fraud misclassification, achieved an accuracy of 94.1% and an AUC-ROC of 0.915, enhancing robustness without altering data distribution. Ultimately, combining SMOTE with cost-sensitive learning provided optimal results—94.7% accuracy and 0.921 AUC-ROC—demonstrating that integrating data augmentation and loss function optimization effectively mitigates imbalance and improves fraud detection performance. The corresponding training loss curves are presented in Figure 2.

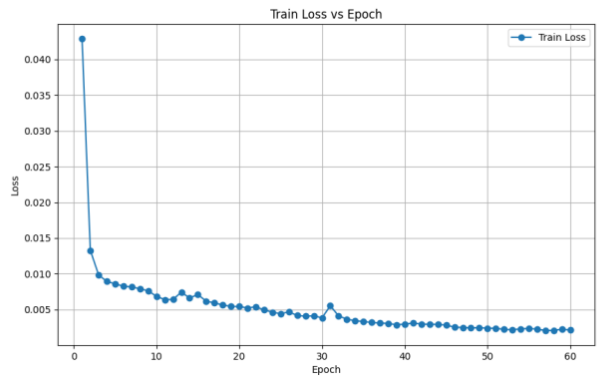

Figure 2. Training loss function decline graph

The experimental results demonstrate a decreasing trend in the loss during the model training process. As illustrated in Figure 2, the initial training phase exhibits a high loss value. However, with the advancement of training, the loss value rapidly decreases and stabilizes after approximately 10 epochs. This indicates that the model swiftly acquires the crucial features during the initial iterations and progressively optimizes its parameters. As the number of epochs increases, the loss gradually decreases and eventually converges after 50 rounds. This suggests that the model has reached the optimal convergence state under the current hyperparameter configuration, and no significant overfitting occurs.

It can be observed from the figure that there is a slight fluctuation in the loss curve at part of the epoch (such as around 20 rounds), which may be due to the dynamic adjustment of the optimizer or the influence of batch gradient update, but the overall downward trend is stable, indicating that the model training is smooth. In addition, the final loss value is low, indicating that the model has a good fitting effect on the fraud detection task. However, the generalization ability of the model cannot be completely judged by training loss alone, and its detection performance should be further evaluated by combining the AUC-ROC and accuracy on the test set.

## IV. Conclusion

This study proposes a credit card fraud detection method based on heterogeneous graph neural networks and validates its effectiveness through experiments on real transaction datasets. By constructing a heterogeneous graph with multiple node types, including users, merchants, and transactions, and incorporating a graph attention mechanism and temporal decay strategy, the model enhances fraud detection capabilities. Experimental results demonstrate that the proposed method outperforms traditional graph neural networks, such as GCN, GAT, and GraphSAGE, in terms of accuracy and OC-ROC. This finding confirms the advantages of heterogeneous graph modeling in fraud detection tasks. Additionally, the combination of SMOTE and cost-sensitive learning mitigates the issue of data imbalance, ensuring high detection performance even when fraudulent transactions are relatively scarce. These findings provide a novel approach to financial risk control and contribute to improving the reliability and intelligence of credit card fraud detection systems.

Despite the promising results, several areas warrant further optimization. This study employs a temporal decay strategy to incorporate time-dependent transaction patterns. However, it does not fully capture long-term dependencies in transaction sequences. Aggregation features based on time series data could be helpful to build fraud detection models. [22] Future work could integrate Transformer or LSTM architectures to enhance temporal modeling and improve the model's generalization over time. Moreover, experimental observations indicate that fraud groups often use forged identities and abnormal transaction patterns to evade detection. Leveraging graph structures to detect organized fraud remains an important research direction. Further research could incorporate multi-modal data, such as user behavior logs, geographic location information, email address attributes, and device fingerprints, to enhance fraud detection accuracy. In practical applications, fraud patterns evolve in response to changes in security policies. New patterns of fraudulent behavior are applied to evade detection systems. Therefore, an adaptive fraud detection framework based on reinforcement learning and real-time learning could be explored to enable the model to dynamically adjust strategies in response to emerging fraud tactics. Addressing these issues will contribute to making credit card fraud detection systems more intelligent and efficient, providing stronger safeguards for financial security.

## References

[1] M. T. Singh, R. K. Prasad, G. R. Michael, et al., "Heterogeneous Graph Auto-Encoder for CreditCard Fraud Detection," arXiv preprint arXiv:2410.08121, 2024.

[2] E. Li, J. Ouyang, S. Xiang, et al., "Relation-aware heterogeneous graph neural network for fraud detection," Proceedings of the Asia-Pacific Web (APWeb) and Web-Age Information Management (WAIM) Joint International Conference on Web and Big Data, pp. 240-255, 2024.

[3] K. Yan, J. Gao and D. Matsypura, "FIW-GNN: A Heterogeneous Graph-Based Learning Model for Credit Card Fraud Detection," Proceedings of the 2023 IEEE 10th International Conference on Data Science and Advanced Analytics (DSAA), pp. 1-10, 2023.

[4] Y. Yao, "Stock Price Prediction Using an Improved Transformer Model: Capturing Temporal Dependencies and Multi-Dimensional Features," Journal of Computer Science and Software Applications, vol. 5, no. 2, 2024.

[5] A. C. Hiremath, A. Arya, L. Sriranga, et al., "Ensemble of Graph Neural Networks for Enhanced Financial Fraud Detection," Proceedings of the 2024 IEEE 9th International Conference for Convergence in Technology (I2CT), pp. 1-8, 2024.

[6] S. Wang and P. S. Yu, "Graph neural networks in anomaly detection," Graph Neural Networks: Foundations, Frontiers, and Applications, pp. 557-578, 2022.

[7] Y. Qi, Q. Lu, S. Dou, X. Sun, M. Li and Y. Li, "Graph Neural Network-Driven Hierarchical Mining for Complex Imbalanced Data," arXiv preprint arXiv:2502.03803, 2025.

[8] J. Wang, "Credit Card Fraud Detection via Hierarchical Multi-Source Data Fusion and Dropout Regularization," Transactions on Computational and Scientific Methods, vol. 5, no. 1, 2025.

[9] P. Feng, "Hybrid BiLSTM-Transformer Model for Identifying Fraudulent Transactions in Financial Systems," Journal of Computer Science and Software Applications, vol. 5, no. 3, 2025.

[10] X. Wang, "Data Mining Framework Leveraging Stable Diffusion: A Unified Approach for Classification and Anomaly Detection," Journal of Computer Technology and Software, vol. 4, no. 1, 2025.

[11] Y. Wang, Z. Xu, Y. Yao, J. Liu and J. Lin, "Leveraging Convolutional Neural Network-Transformer Synergy for Predictive Modeling in Risk-Based Applications," arXiv preprint arXiv:2412.18222, 2024.

[12] T. Zhou, Z. Xu and J. Du, "Efficient Market Signal Prediction for Blockchain HFT with Temporal Convolutional Networks," Transactions on Computational and Scientific Methods, vol. 5, no. 2, 2025.

[13] M. Jiang, Z. Xu and Z. Lin, "Dynamic Risk Control and Asset Allocation Using Q-Learning in Financial Markets," Transactions on Computational and Scientific Methods, vol. 4, no. 12, 2024.

[14] X. Sun, Y. Yao, X. Wang, P. Li and X. Li, "AI-Driven Health Monitoring of Distributed Computing Architecture: Insights from XGBoost and SHAP," arXiv preprint arXiv:2501.14745, 2024.

[15] G. Huang, Z. Xu, Z. Lin, X. Guo and M. Jiang, "Artificial Intelligence-Driven Risk Assessment and Control in Financial Derivatives: Exploring Deep Learning and Ensemble Models," Transactions on Computational and Scientific Methods, vol. 4, no. 12, 2024.

[16] X. Wang, L. Xiangfeng, X. Wang, et al., "Homophilic and Heterophilic-Aware Sparse Graph Transformer for Financial Fraud Detection," Proceedings of the 2024 International Joint Conference on Neural Networks (IJCNN), pp. 1-8, 2024.

[17] J. Liu, "Multimodal Data-Driven Factor Models for Stock Market Forecasting," Journal of Computer Technology and Software, vol. 4, no. 2, 2025.

[18] S. Xiang, M. Zhu, D. Cheng, et al., "Semi-supervised credit card fraud detection via attribute-driven graph representation," Proceedings of the AAAI Conference on Artificial Intelligence, vol. 37, no. 12, pp. 14557-14565, 2023.

[19] J. Hu, T. An, Z. Yu, J. Du and Y. Luo, "Contrastive Learning for Cold Start Recommendation with Adaptive Feature Fusion," arXiv preprint arXiv:2502.03664, 2025.

[20] G. Zhang, Z. Li, J. Huang, et al., "efraudcom: An e-commerce fraud detection system via competitive graph neural networks," ACM Transactions on Information Systems (TOIS), vol. 40, no. 3, pp. 1-29, 2022.

[21] L. Ren, R. Hu, D. Li, et al., "Dynamic graph neural network-based fraud detectors against collaborative fraudsters," Knowledge-Based Systems, vol. 278, pp. 110888, 2023.

[22] F. Xiao, Y. Wu, M. Zhang, G. Chen and B. C. Ooi, "MINT: Detecting Fraudulent Behaviors from Time-Series Relational Data," Proceedings of the VLDB Endowment, vol. 16, no. 12, pp. 3610–3623, 2023. DOI: 10.14778/3611540.3611551.